# Tractable Bayesian Learning of Tree Belief Networks

**Marina Meilă**
Carnegie Mellon University
mmp@cs.cmu.edu

**Tommi Jaakkola**
Massachusetts Institute of Technology
tommi@ai.mit.edu

## Abstract

In this paper we present *decomposable priors*, a family of priors over structure and parameters of tree belief nets for which Bayesian learning with complete observations is tractable, in the sense that the posterior is also decomposable and can be completely determined analytically in polynomial time. This follows from two main results: First, we show that factored distributions over spanning trees in a graph can be integrated in closed form. Second, we examine priors over tree parameters and show that a set of assumptions similar to (Heckerman and al., 1995) constrain the tree parameter priors to be a compactly parametrized product of Dirichlet distributions. Besides allowing for exact Bayesian learning, these results permit us to formulate a new class of tractable latent variable models in which the likelihood of a data point is computed through an ensemble average over tree structures.

## 1 Introduction

In the framework of graphical models, tree distributions stand out by their special computational advantages. Inference and sampling from a tree are linear in the number of variables $n$. While it is known that for many classes of graphical models, as for example junction trees with cliquewidth > 2, the problem of learning the optimal structure is NP-hard, for trees this problem is solvable in only quadratic time. The latter result is due to [Chow and Liu, 1968] who present an algorithm for finding the structure and parameters of the tree that best fits a given distribution in the Maximum Likelihood (ML) framework. This algorithm was generalized to Maximum A-Posteriori (MAP) learning [Meilă-Predoviciu, 1999, Heckerman et al., 1995].

In this paper we present another remarkable property of tree graphical models: the fact that Bayesian learning for a certain class of priors, called *decomposable*[1] priors, is also tractable. Essentially, decomposable priors are priors that can be represented as a product of factors corresponding to the edges of the tree. We show that if the prior is decomposable and we have a data set consisting of $N$ complete i.i.d. observations, then the posterior distribution over all tree structures and parameters is also decomposable, is expressible with a quadratic number of parameters that can be computed exactly from data in $\mathcal{O}(n^3 + n^2 N)$ operations. Evaluating the posterior for a given tree takes then $\mathcal{O}(n)$ time. The first two results come from the fact that, with the standard assumptions of likelihood equivalence, parameter independence and parameter modularity, the prior for tree parameters is constrained to be a product of Dirichlet distributions whose parameters satisfy a set of consistency relations. The last result, i.e. the possibility of computing the posterior exactly, is a consequence of the fact that a factored distribution over tree structures can be integrated exactly, using a theorem from combinatorics called *the Matrix tree theorem*.

---

[1] The term decomposable prior will refer here to a prior over a family of graphical models. It should not be confused with a *decomposable model* which is a distribution over $V$.

The paper starts by defining tree distributions and the problem of Bayesian learning in section 2; it presents decomposable priors over tree structures and parameters in sections 3 and 4; the pieces of the puzzle are put together in section 5 where Bayesian learning is described; the next section, 6 exploits a different set of possibilities opened by our tractability results: it defines a new model, *ensembles of trees*, and shows that it can be learned by gradient ascent in the ML framework; section 7 contains the final remarks.

## 2 Tree distributions and the Bayesian learning problem

In this section we introduce the tree model and the notation that will be used throughout the paper. Let $V = \{1, \ldots, n\}$ denote the set of variables of interest. Let $r_v$ be the number of values of variable $v \in V$, $r_{MAX} = \max r_v$, $x_v$ a particular value of $v$, and $x$ an assignment to all the variables in $V$.

According to the graphical model paradigm, each variable is viewed as a vertex of a graph. We shall call a graph that has no cycles a *tree*[2] and shall denote by $E$ its edge set. If the tree is connected, e.g. it *spans* all the nodes in $V$, it is called a *spanning tree*.

Now we define a probability distribution $T$ that is *conformal* with a tree. Let us denote by $T_{uv}$ and $T_v$ the marginals of $T$:

$$T_{uv}(x_u, x_v) = \sum_{x: u=x_u, v=x_v} T(x)$$
$$T_v(x_v) = \sum_{x: v=x_v} T(x).$$

Let $\deg v$ be the *degree* of vertex $v$, e.g. the number of edges incident to $v \in V$. Then, the distribution $T$ is conformal with the tree $(V, E)$ if it can be factored as:

$$T(x) = \frac{\prod_{uv \in E} T_{uv}(x_u, x_v)}{\prod_{v \in V} T_v(x_v)^{\deg v - 1}} \quad (1)$$

---
[2]In the graph theory literature, our definition corresponds to a *forest*. The connected components of a forest are called trees.

The distribution itself will be called a tree when no confusion is possible. An equivalent representation for $T$ in terms of conditional probabilities is

$$T(x) = \prod_{v \in V} T_{v|\text{pa}(v)}(x_v | x_{\text{pa}(v)}) \quad (2)$$

where $\text{pa}(v)$ represents the parent of $v$ in the thus directed tree or the empty set if $v$ is the root of a connected component. The form (2) can be obtained from (1) by choosing an arbitrary root in each connected component and recursively substituting $\frac{T_{v,\text{pa}(v)}}{T_{\text{pa}(v)}}$ by $T_{v|\text{pa}(v)}$ starting from the root. We denote such a *directed* tree structure by $\overline{E}$. The directed tree representation has the advantage of having independent parameters. The total number of free parameters in either representation is

$$\sum_{uv \in E} r_u r_v - \sum_{v \in V} (\deg v - 1) r_v.$$

In the forthcoming we shall use both representations. Which representation we consider will be clear from the context in all cases of relevance.

We now turn to the problem of learning trees in the Bayesian framework. In this framework, one assumes a prior $P_0(T)$ over the set $\mathcal{T}_V$ of all tree distributions defined on the domain $V$. Learning from a dataset of complete and independently generated observations $\mathcal{D} = \{x^1, x^2, \ldots x^N\}$ means finding the *posterior* distribution $P(T|\mathcal{D})$ over the set of models $\mathcal{T}_V$. The solution to this problem is given by the well known Bayes' formula

$$P(T|\mathcal{D}) \propto P_0(T) \prod_{t=1}^{N} T(x^t) \quad (3)$$

Practically however, Bayesian learning poses a number of significant challenges. First, one needs to define a distribution over the space of all models to play the role of the prior. Such a distribution is composed of a discrete distribution over the set of tree structures $P_0(E)$ and a probability density over the continuous set of tree parameters $P_0(\theta|E)$. Here $\theta$ consists of all the parameters of $T$ in some representation.

$$P_0(T) = P_0(E) P_0(\theta | E) \quad (4)$$

The discrete space of all tree structures over $V$ has a super-exponential number of trees (order $n^{n-2}$)

which makes defining a distribution over such a space a non-trivial task. Moreover, the second factor in the above formula requires us to define a prior distribution for the tree parameters for each possible structure $E$. Thus, the first practical requirement is to have a tractable representation for the prior. Even with a tractable representation, the explicit computation of the posterior $P(T|\mathcal{D})$ is usually intractable due to the difficulty of computing the normalization constant in (3). Therefore common practices in Bayesian learning are Maximum A-Posteriori (MAP) estimation and approximations of the posterior around its peaks. An exception from this situation are the so-called *conjugate priors*. If a given (graphical) model has a family of conjugate priors $\mathcal{P}$ then for $P_0 \in \mathcal{P}$ the posterior is also in $\mathcal{P}$. The property of having conjugate priors is characteristic of the exponential family of distributions [DeGroot, 1975]. In this paper we set out to find the conjugate prior for the family of tree models $\mathcal{T}_V$.

According to (4), to define a prior over $\mathcal{T}_V$ one needs to define a prior over tree structures and a prior for tree parameters, given the structure. While it is not hard to see that for a fixed structure $E$ a tree distribution over discrete variables is an exponential model and thus has conjugate priors, realizing the same fact when $E$ also varies is by far less obvious and constitutes the main contribution of this paper. In the next section we establish the core theorem that allows us to do so.

## 3 Decomposable priors over tree structures

A *decomposable* distribution $P$ over spanning tree structures $E$ depends on a set of parameters $\beta_{uv} = \beta_{vu} \geq 0$; $\beta_{vv} = 0$, $u, v \in V$ by

$$P(E) = \frac{1}{Z} \prod_{uv \in E} \beta_{uv}. \tag{5}$$

In the above, $Z$ is the normalization constant

$$Z = \sum_E \prod_{uv \in E} \beta_{uv}. \tag{6}$$

Note that in the distribution (6), each parameter $\beta_{uv}$ can be interpreted as the weight of edge $uv$, and the probability of a structure $E$ is the product of the weights of all edges in $E$. Although this distribution is expressible in a product form, it does not imply that the edges' occurrences in $E$ are independent, since the set $E$ as a whole is constrained to be a tree structure.

This prior is simple and compactly parametrized, but to be completely defined one needs to evaluate the normalization constant $Z$. Using formula (6) is intractable, but the following theorems develop a practical and exact method for it.

**Theorem 1 (Matrix Tree Theorem)**
[West, 1996] *Let $G = (V, E)$ be a multigraph and denote by $a_{uv} = a_{vu}$ the number of undirected edges between vertices $u$ and $v$. Then the number of all spanning trees of $G$ is given by $|A_{uv}|(-1)^{u+v}$ the value of the determinant obtained from the following matrix by removing row $u$ and column $v$[3].*

$$A = \begin{bmatrix} \deg v_1 & -a_{12} & -a_{13} & \cdots & -a_{1,n} \\ -a_{21} & \deg v_2 & -a_{23} & \cdots & -a_{2,n} \\ \cdots & \cdots & \cdots & \cdots & \cdots \\ -a_{n,1} & -a_{n,2} & -a_{n,3} & \cdots & \deg v_n \end{bmatrix} \tag{7}$$

In the following, we shall use the simplifying notation below to refer to a set of real values each corresponding to a pair of variables in $V$

$$a = \{a_{uv}, u, v \in V, u \neq v\} \tag{8}$$

In addition, $a \geq 0$ will mean that $a_{uv} \geq 0$, $a_{uv} \in a$ and $ab$ will denote $\{a_{uv} b_{uv}, u, v \in V, u \neq v\}$ for $a, b$ defined as above. By extending the Matrix Tree theorem to continuous valued $A$ and letting the weights $\beta$ play the role of $a$ in (7), one can prove

**Theorem 2** [Jaakkola et al., 2000] *Let $P(E)$ be a distribution over spanning tree structures defined by (5,6). Then the normalization constant $Z$ is equal to $|Q(\beta)|$ with $Q(\beta)$ being the first $(n-1)$ lines and columns of the matrix $\overline{Q}(\beta)$ given by:*

$$\overline{Q}_{uv}(\beta) = \overline{Q}_{vu}(\beta) = \begin{cases} -\beta_{uv} & 1 \leq u < v \leq n \\ \sum_{v'=1}^{n} \beta_{v'v} & 1 \leq u = v \leq n \end{cases} \tag{9}$$

---

[3] Note that $A$ as a whole is a singular matrix.

This shows that summing over the distribution of all spanning trees, when this distribution factors according to the trees' edges, can be done in closed form by computing the value of an order $n - 1$ determinant, operation that involves $\mathcal{O}(n^3)$ operations. The proof of theorem 2 as well as the other proofs appear in [Meilă and Jaakkola, 2000].

In the following it will be useful to think of $\overline{Q}(\beta)$ and $Q(\beta)$ as functions mapping a set of parameters $\beta$ each corresponding to a pair of variables in $V$ into a matrix the ways described by theorem 2.

**The support graph.** The factored form of the decomposable distribution makes it easy to test whether a given structure has non-zero probability. If all the $\beta$ parameters are strictly positive, then every tree structure is possible. Otherwise, the structures that will never appear are the structures containing one or more zero-weight edges. We denote by $E^{sup}$ the set of edges $uv$ for which $\beta_{uv} > 0$. The graph $G^{sup} = (V, E^{sup})$ is called *the support graph* of $P(E)$. If enough edges have zero weights, then $G^{sup}$ may be disconnected. In the following we shall assume that the support graph is connected, leaving the discussion of the general case for [Meilă and Jaakkola, 2000].

In the remainder of this section we develop a number of consequences of theorem 2.

**Computing averages under a decomposable distribution** A decomposable distribution is a (curved) [Murray and Rice, 1993] exponential model and $\ln Z$ represents its *cumulant generating function* or *partition function*; many quantities of interest, like averages under $P(E)$ can be expressed as derivatives of the partition function. The next series of results exemplifies these possibilities.

**Lemma 3** [Jaakkola et al., 2000] *Let $Z$ be given by equation (6) with $\beta \geq 0$, $Q(\beta)$ be given by theorem 2, $Q^{-1}$ be the inverse of $Q$ and $M(\beta)$ be a symmetric matrix with 0 diagonal defined by*

$$M_{uv} = (Q^{-1})_{uu} + (Q^{-1})_{vv} - 2(Q^{-1})_{uv}, \quad u, v < n$$
$$M_{nv} = M_{vn} = (Q^{-1})_{vv}, \quad v < n \qquad (10)$$
$$M_{vv} = 0$$

*Then the partial derivative of $Z$ with respect to $\beta_{uv}$ is*

$$\frac{\partial Z}{\partial \beta_{uv}} = M_{uv}(\beta)|Q(\beta)|. \qquad (11)$$

We shall denote by $<f>_P$ the average of a function $f$ under distribution $P$. The following lemma states a useful fact about averages of additive functions. An *additive* function $f(E)$ satisfies

$$f(E) = \sum_{uv \in E} f_{uv} \qquad (12)$$

**Lemma 4** [Jaakkola et al., 2000] *Let $P(E), Q$ and $M$ be given by (5), theorem 2 and (10) respectively and $f$ be an additive function of the structure $E$. Then the average of $f$ under $P$ is*

$$<f(E)>_P = \sum_E f(E)P(E) \qquad (13)$$
$$= \sum_{u<v} f_{uv}\beta_{uv}M_{uv}(\beta) \qquad (14)$$
$$= \operatorname{trace} Q(f\beta)Q^{-1}(\beta) \qquad (15)$$

In (15), $f$ is an overloaded notation representing the set $\{f_{uv}, u, v \in V\}$ in the sense of (8). A similar but more obvious result holds for functions $g(E)$ that are *multiplicative*, i.e. $g(E) = \prod_{uv \in E} g_{uv}$. For such functions we obtain

$$<g(E)>_P = \frac{|Q(\beta g)|}{|Q(\beta)|} \qquad (16)$$

## 4 Decomposable priors over tree parameters

Now we examine priors over tree parameters, with the goal of finding conditions under which the priors can be tractably represented. The assumptions we make are similar to those of [Heckerman et al., 1995] (called HGC in the forthcoming) and so will be some of the results. In addition, we will show that in the case of trees these assumptions are also sufficient for tractable representation and learning.

In the following, without loss of generality, we will consider that both the directed and the undirected tree representation are in the *probability table* parametrizations, and we denote respectively

$$\theta_v(j) = T_v(j) \qquad (17)$$
$$\theta_{uv}(ij) = T_{uv}(ij) \qquad (18)$$
$$\theta_{u|v}(i|j) = T_{u|v}(i|j) \qquad (19)$$

and $\theta_E = \{\theta_{uv}(ij), uv \in E, i = 1,\ldots r_u, j = 1,\ldots r_v\}$, $\theta_{\overline{E}} = \{\theta_{u|v}(ij), \overline{vu} \in \overline{E}, i = 1,\ldots r_u, j = 1,\ldots r_v\} \cup \{\theta_v(j), v\,\text{root}, j = 1\ldots r_v\}$. We will assume by convention that if $v \in V$ has no parent then $\theta_{v|\text{pa}(v)} = \theta_v$ and that $\text{pa}(v)$ takes one value only.

First let us keep the distribution $T$ fixed. As shown in section 2 this distribution can be represented either by (1) or by (2), the latter representation having a distinct form for each possible choice of the root(s). These representations however will assign exactly the same probability $T(x)$ to an observation $x$, so there is no way to distinguish between them from the point of view of the data. Thus we shall require that the corresponding parameter sets are also the same from the point of view of the prior. This leads to the assumption of *Likelihood equivalence*:

**Assumption 1 (Likelihood equivalence)**
*Let $T$ be a tree distribution having structure $E$, $\overline{E}$ a directed tree structure obtained from $E$ and $\theta_E, \theta_{\overline{E}}$ the respective parameters of $T$. Denote by $|\frac{\partial \theta_{\overline{E}}}{\partial \theta_E}|$ the magnitude of the Jacobian of the transformation $\theta_E \to \theta_{\overline{E}}$. Then $P_0(\theta_{\overline{E}}(\theta_E)|\overline{E})|\frac{\partial \theta_{\overline{E}}}{\partial \theta_E}| = P_0(\theta_E|E)$.*

This assumption states that in all possible parametrizations consistent with a given structure $E$ the prior will assign the same probability mass to any given (measurable) subset in parameter space. Thus, the prior treats likelihood equivalent parametrizations as indistinguishable.

Likelihood equivalence has somewhat compressed the space that we have to define $P_0$ on, but it still leaves us with the task of assigning a separate prior for each (undirected) tree structure. We now transform the problem into one of assigning a prior for each of the possible tree edges by making the following additional assumptions:

**Assumption 2 (Parameter independence)**
*For any structure $\overline{E}$ and any $\overline{vu} \in \overline{E}, j, j' = 1,\ldots r_v$ the parameter vectors $\theta_{u|v}(.|j)$ and $\theta_{u|v}(.|j')$ are independent under $P_0$. The parameters $\theta_{u|v}(.|j)$ are also independent under $(P_0)$ of the parameter sets $\theta_{u'|v'}(.|j')$ corresponding to any other edge in $\overline{E}$.*

**Assumption 3 (Parameter modularity)**
*The prior $P_0(\theta_{u|v}|\overline{E})$ is the same for all structures $\overline{E}$ that contain the edge $\overline{vu}$.*

In other words, parameter independence states that the prior over parameters factors into a product over the edges; by stating in addition that the prior for an edge is the same for all tree structures that contain that edge, we have effectively removed the dependence on $E$ from the parameters prior. From now on, we will write $P_0(\theta_E)$, $P_0(\theta_{\overline{E}})$ instead of $P_0(\theta_E|E)$ and $P_0(\theta_{\overline{E}}|\overline{E})$ respectively. We shall call a prior $P_0$ satisfying assumptions 1, 2 and 3 a *decomposable prior* for tree parameters. If both $P_0(E)$ and $P_0(\theta)$ are decomposable, the resulting prior over tree distribution is also called *decomposable*. As we shall see next, the same assumptions also constrain the functional form the prior can have. Again we assume that $G^{sup}$ is connected.

**Assumption 4 (Connectivity)** *The support graph of $P_0(E)$ is connected.*

**Theorem 5** *Let $P(T) = P(E)P(\theta_E)$ be a decomposable distribution over tree parameters, for which the support graph of $P(E)$ is connected and $P(\theta_E) > 0$ for $\theta_E > 0$. Then for any tree $T$ in any directed representation $\overline{E}, \theta_{\overline{E}}$:*

$$P(\theta_{\overline{E}}|\overline{E}) = \prod_{v \in V} P(\theta_{v|\text{pa}(v)}) \quad (20)$$

$$P(\theta_{v|u}) = \prod_{i=1}^{r_u} D(\theta_{v|u}(.|i); N'_{vu}(.i)) \quad (21)$$

*where $D$ is the Dirichlet distribution and $N'_{vu}(ij) > 0$ are its parameters. The numbers $N'_{uv}(ij) = N'_{vu}(ji)$ are defined for all edges $uv$ with $\beta_{uv} > 0$ and satisfy*

$$\sum_{i=1}^{r_u} N'_{uv}(ij) = N'_v(j) \quad (22)$$

$$\sum_{j=1}^{r_v} N'_v(j) = N' \quad (23)$$

The Dirichlet prior is defined over the parameter space $\theta_1,\ldots\theta_r$, $(\sum_j \theta_j = 1, \theta_j > 0, j = 1,\ldots r)$ of a distribution over a discrete set by

$$D(\theta_1,\ldots\theta_r; N'_1,\ldots N'_r) = \frac{1}{Z_D} \prod_{j=1}^{r} \theta_j^{N'_j - 1} \quad (24)$$

The numbers $N'_1, \ldots N'_r > 0$ are the hyperparameters of the Dirichlet prior, their sum is denoted by $N'$. The normalization constant $Z_D$ has the form

$$Z_D = \frac{\prod_{j=1}^{r} \Gamma(N'_j)}{\Gamma(N')} \qquad (25)$$

with $\Gamma$ being the Euler function $\Gamma(p) = \int_0^\infty x^{p-1} e^{-x} dx$.

The above line of reasoning parallels the one in HGC. The assumptions 1 – 3 are the specialization for tree structures of their homonyms in HGC. But unlike the case of general Bayes nets, where the prior is in general specified by an exponential number of parameters, in the case of tree graphical models the prior can be specified by a set of only $\mathcal{O}(n^2 r_{MAX}^2)$ "pairwise marginal counts" $N'_{uv}(ij)$. This is possible because in the space of tree structures the likelihood equivalence classes can be explicitly represented and the number of possible parents for a variable is no larger than one. Therefore, not only the tree belief net itself, but also any decomposable distribution over trees can be completely defined in terms of pairwise interactions[4].

The same properties allow us to replace a fourth assumption made by HGC, namely structure possibility, with the weaker assumption 4. Note that if all $\beta_{uv} > 0$, then all tree structures are possible, $G^{sup}$ is connected and our theorem 5 is an exact rewrite of the similar result in HGC. This last assumption is not essential for our results. In [Meilă and Jaakkola, 2000] we give a general formulation of the above theorem that dispenses with the connectivity assumption.

To summarize, starting with the assumptions 1–3 and aiming mainly at obtaining a tractable and consistent prior representation, we have arrived at the conclusion that the prior has to be a product of Dirichlet distributions. This demonstrates that our initial requirement is a drastic one; the restrictions on the prior should be understood as restrictions on the type of prior information about the model we are allowed to have. A Dirichlet distribution means essentially that we have only knowledge about the values of the parameters' means. This issue is further developed in HGC to which we refer the reader. On the computational side however, the advantage is enormous, since with the Dirichlet distribution Bayesian learning is possible in closed form. The next section will exploit exactly this property to find the posterior over tree distributions.

## 5 Bayesian learning with decomposable priors

From equations (1,2) we know that the likelihood can be written as a product over tree edges. Theorem 5 proves the same thing about the decomposable prior. It follows then that the posterior $P(T|\mathcal{D})$ in equation (3) can also be factored over the edges of $T$. We shall see that in addition $P(T|\mathcal{D})$ is decomposable and the normalization constant $P(\mathcal{D}) = Z_\mathcal{D}$ can be computed tractably.

We shall use the following important property of a Dirichlet distribution: Assume a discrete variable $z$ that takes values $1 \ldots r$ with probabilities $\theta = (\theta_1, \ldots \theta_r)$, a prior for $\theta$ that is Dirichlet with parameters $N'(1), \ldots N'(r)$ and a set $\mathcal{D}_z$ of $N$ independent observations for $z$, such that the value $j$ appears $N(j)$ times in $\mathcal{D}_z$. Then, the posterior of the parameters $\theta$ is (see e.g. [DeGroot, 1975]) also Dirichlet with parameters $N'(j) + N(j)$.

This result applies immediately to the posterior of a tree. Let us denote by $N_{uv}(ij)$ and $N_v(j)$ the *sufficient statistics* of the sample $\mathcal{D}$, i.e. the number of times $u = i, v = j$ and respectively $v = j$ in $\mathcal{D}$. Then, from the above and theorem 5 we obtain

$$P(T|\mathcal{D}) \propto \prod_{uv \in E} \beta_{uv} \prod_{\overline{uv} \in \overline{E}} \prod_{i=1}^{r_u} D(\theta_{v|u}(.|i); N_{uv}(i.) + N'_{uv}(i.)) \qquad (26)$$

Hence, $P(T|\mathcal{D})$ is also a decomposable distribution over tree structures, and its parameters are available directly from the parameters of the prior and the sufficient statistics of the sample. It remains to show how to compute the normalization constant in (3). For this, we will first keep the structure $E$ fixed and integrate over the parameters $\theta_{\overline{E}}$ in some directed structure $\overline{E}$ obtained from $E$.

$$\int P(T|\mathcal{D}) d\theta_{\overline{E}} = \qquad (27)$$

---

[4]It is important to note that the parameters $N'_{uv}(ij)$ cannot be set arbitrarily. They have to be proportional to the marginals of some distribution over $V$.

$$\propto \prod_{uv \in E} \beta_{uv} \prod_{\overline{uv} \in \overline{E}} \prod_{i=1}^{r_u} \int D(\theta_{v|u}(.|i); N_{uv}(i.) + N'_{uv}(i.))d\theta_{v|u}$$

$$= \frac{\Gamma(N')}{\Gamma(N'+N)} \prod_{v \in V} \prod_{j=1}^{r_v} \frac{\Gamma(N'_v(j))}{\Gamma(N_v(j) + N'_v(j))}$$

$$\cdot \prod_{uv \in E} \beta_{uv} \prod_{i=1}^{r_u} \prod_{j=1}^{r_v} \frac{\Gamma(N_{uv}(ij) + N'_{uv}(ij))}{\Gamma(N'_{uv}(ij))}$$

This quantity represents the marginal posterior $P(E|\mathcal{D})$; as required by likelihood equivalence, it is the same no matter how $\overline{E}$ is obtained from $E$. Note also that $P(E|\mathcal{D})$ decomposes into a product over the edges in $E$ preceded by a factor independent of $E$. We define the weights $W_{uv}$ as

$$W_{uv} = \prod_{i=1}^{r_u} \prod_{j=1}^{r_v} \frac{\Gamma(N_{uv}(ij) + N'_{uv}(ij))}{\Gamma(N'_{uv}(ij))} \quad (28)$$

Now we can apply theorem 2 to the weights $\beta W$ obtaining

$$Z_\mathcal{D} = \sum_E P(E|\mathcal{D}) = |Q(\beta W)| \quad (29)$$

This completely defines the posterior distribution $P(T|\mathcal{D})$. The posterior probability of any tree distribution $T$ can be now computed analytically based on equations (26) and (29) while (27) and (29) give the posterior of any tree structure $E$. Note that the weights $W_{uv}$ are never 0, so that the support graph of the posterior distribution coincides with the $G^{sup}$ of the prior.

To compute the posterior representation from the data set we need $\mathcal{O}(n^2 r^2_{MAX} N)$ operations to obtain the sufficient statistics and to evaluate the edge weights $W_{uv}$ and an additional $\mathcal{O}(n^3)$ to evaluate the normalization constant $Z_\mathcal{D}$ for a total of $\mathcal{O}(n^2 r^2_{MAX} N + n^3)$ operations. Computing the posterior of a tree (or tree structure) is now $\mathcal{O}(n r_{MAX})$.

Furthermore, to perform Bayesian averaging in computing the probability of a new data point $x$ one has to evaluate

$$P(x|\mathcal{D}) = \int T(x) P(T|\mathcal{D}) dT \quad (30)$$

Just as before, we can first integrate the above expression over the parameters for a fixed structure $E$ and them perform a summation over structures. The former step yields

$$\int \prod_{v \in V} T_{v|\text{pa}(v)}(x_v|x_{\text{pa}(v)}) P(\theta_{\overline{E}}|\mathcal{D}) d\theta_{\overline{E}} = \quad (31)$$

$$= \underbrace{\frac{1}{N'+N} \prod_{v \in V}[N'_{\text{pa}(v)}(x_{\text{pa}(v)}) + N_{\text{pa}(v)}(x_{\text{pa}(v)})]}_{w_0(x)} \cdot$$

$$\prod_{uv \in E} \underbrace{\frac{N'_{vu}(x_v x_u) + N_{v,u}(x_v x_u)}{[N'_u(x_u) + N_u(x_u)][N'_v(x_v) + N_v(x_v)]}}_{w_{uv}(x)}$$

Again, we note that the result includes a structure independent factor $w_0(x)$ and a product of factors corresponding to the tree edges $w_{uv}(x)$. Also, we note that the final result is invariant to the particular orientation $\overline{E}$ of $E$. Summing now over tree structures is a mere exercise; we get

$$P(x|\mathcal{D}) = \sum_E \frac{w_0(x)}{Z_\mathcal{D}} \prod_{uv \in E} \beta_{uv} w_{uv}(x)$$

$$= \frac{w_0(x)|Q(\beta w(x))|}{|Q(\beta W)|} \quad (32)$$

The averaging involves computing the edge weights $w(x)$ and evaluating a determinant, so that the total computation is $\mathcal{O}(n^3)$, a relatively large value compared to the $\mathcal{O}(n)$ demands of the ML and MAP tree likelihood.

The result generalizes readily to the Bayesian averaging of the probability of a set of more than one independent observations.

## 6 Ensembles of trees

In this section we consider a new probability model, called *ensembles of trees* that naturally extends the tree graphical model. To best describe this model, imagine that a tree distribution is defined in two steps: first a set of parameters $\theta$ and second the structure $E$. Because $E$ is not known at the time when we choose $\theta$, we need to specify a parameter set that is sufficiently large, so that for any $E$ we can afterwards extract from $\theta$ the actual set of parameters $\theta_E$. This can be done easily via the same idea that allowed us to define a decomposable prior in section 4; we choose $\theta = \{\theta_{uv}(ij), \ u, v \in V, i = 1, \ldots r_u, j =$

$1, \ldots r_v\} \cup \{\theta_v(j), v \in V, j = 1, \ldots r_v\}$ such that

$$\sum_{i=1}^{r_u} \theta_{uv}(ij) = \theta_v(j) \; \forall \, u \in V \quad (33)$$

$$\sum_{j=1}^{r_v} \theta_v(j) = 1 \; \forall v \in V$$

Now, changing the notation of equation (1) to emphasize the dependence on $\theta$ and $E$, and rearranging the factors, we write the tree distribution as

$$T(x|\theta, E) = \prod_{uv \in E} \frac{\theta_{uv}(x_u, x_v)}{\theta_u(x_u)\theta_v(x_v)} \prod_{v \in V} \theta_v(x_v) \quad (34)$$

The ensemble of trees $R(x)$ is a weighted average of all the possible tree distributions sharing the same parameters $\theta$. To ensure tractability, the weights will represent a decomposable distribution over structures as in (5). Again, we assume that only spanning trees are possible.

$$R(x) = \sum_E P(E) T(x|\theta, E) \quad (35)$$

If we use the notations

$$w_{uv}(x) = \frac{\theta_{uv}(x_u, x_v)}{\theta_u(x_u)\theta_v(x_v)} \quad (36)$$

$$w_0(x) = \prod_{v \in V} \theta_v(x_v) \quad (37)$$

for the edge dependent and respectively edge independent factors in (34) then, by theorem 2, $R(x)$ has an alternative, tractable form

$$R(x) = w_0(x) \frac{|Q(w\beta)|}{|Q(\beta)|} \quad (38)$$

The ensemble of trees can be seen as a mixture model whose components are the trees over $V$ parametrized by $\theta$. The weighted averaging corresponds then to the presence of a hidden variable $z$ taking as many values as there are structures, each with probability $Pr[z = E] = P(E)$. Therefore, the generalized EM algorithm [Dempster et al., 1977] can be considered as a possibility for learning the parameters. We shall not pursue this issue here for lack of space, but we will mention the following: the E step of the algorithm is tractable and straightforward given equation (35); the M-step however cannot be performed exactly and it is not known if the expression to be maximized has a unique local maximum.

But if we assume a set of complete observations $\mathcal{D}$ as before, the likelihood of this data set, denoted by $R(\mathcal{D})$, can be optimized w.r.t. the parameters $\theta$ and $\beta$ by gradient ascent. We shall denote by $M_{uv}(\beta)$ and $M_{uv}(\beta w(x^t))$ respectively the values in equation (10) that correspond to $Q(\beta)$ and $Q(\beta w(x^t))$. Using lemma 3 we obtain

$$\frac{\partial \log R(\mathcal{D})}{\partial \beta_{uv}} = \sum_{t=1}^{N} \frac{w_{uv}(x^t) M_{uv}(x^t)}{|Q(\beta w(x^t))|} - N M_{uv}(\beta) \quad (39)$$

$$\frac{\partial \log R(\mathcal{D})}{\partial \theta_{uv}(ij)} = \frac{\beta_{uv}}{\theta_u(i)\theta_v(j)} \sum_{t: x_u^t = i, x_v^t = j} M_{uv}(x^t) \quad (40)$$

$$\frac{\partial \log R(\mathcal{D})}{\partial \theta_v(j)} = \frac{1}{\theta_v(j)} \sum_{t: x_v^t = j} [1 - \sum_{v' \in V} w_{vv'}(x) M_{vv'}(x^t)]$$

(41)

Note that the parameters $\theta$ need to satisfy (33) and therefore we will need to perform a constrained maximization of $R(\mathcal{D})$ using e.g. Lagrange multipliers and that this method will converge to only a local optimum of the log-likelihood.

## 7  Discussion

This paper has presented decomposable priors, a class of priors over tree structures and parameters that makes exact Bayesian learning tractable. A decomposable prior is expressed as a product of factors corresponding to the tree's edges. The same edge contributes the same amount in every tree structure that includes it. This property allows (1) representing the prior by order $n^2$ parameters and (2) using the Matrix tree theorem to integrate the prior in closed form.

It is remarkable that for trees, the standard assumptions of HGC are sufficient to ensure tractability. In fact, these assumptions are no stronger than the assumptions of *functional independence* implicit in the original Chow and Liu algorithm [Chow and Liu, 1968, Meilă-Predoviciu, 1999].

But it is worth highlighting again that these assumptions are restrictive, in the sense of drastically limiting the type of prior knowledge that can be used efficiently in the Bayesian learning of trees. Knowledge that violates assumptions 1–3 is e.g. knowledge that two edges are more likely to appear simultaneously than separately in a tree structure, or knowledge that two edges have the same parameters. This problem is not specific to trees, but to Bayes nets in general. Therefore, a worthwhile area of future research is discovering tractable methods to deal with such type of knowledge in the case of tree structures or in the case of general Bayes nets.

We have also introduced ensembles of trees as a tractable extension to the tree model. Ensembles of trees can be learned in the ML framework. Exploring the properties of the new model and of the learning algorithm itself are areas of continuing research.